\documentclass[letterpaper, 10 pt, conference]{ieeeconf}  

\IEEEoverridecommandlockouts                              

\usepackage{graphics} 
\usepackage{epsfig} 
\usepackage{mathptmx} 
\usepackage{times} 
\usepackage{amsmath} 
\usepackage{amssymb}  
\usepackage{lipsum}
\usepackage{color}

\title{\LARGE \bf
Real-time 3D Shape Instantiation from Single Fluoroscopy Projection for Fenestrated Stent Graft Deployment
}

\author{Xiao-Yun Zhou$^{1}$, Jianyu Lin$^{1}$, Celia Riga$^{2}$, Guang-Zhong Yang$^{1}$ and Su-Lin Lee$^{1}$
\thanks{$^{1}$Xiao-Yun Zhou, Jianyu Lin, Guang-Zhong Yang, Su-Lin Lee are with the Hamlyn Centre for Robotic Surgery, Imperial College London, UK
        {\tt\small xiaoyun.zhou14@imperial.ac.uk}}%
\thanks{$^{2}$Celia Riga is with the Regional Vascular Unit, St Mary's Hospital, London, UK and the Academic Division of Surgery, Imperial College London, UK}%
}

\begin{document}

 \maketitle
\thispagestyle{empty}
\pagestyle{empty}

\begin{abstract}
Robot-assisted deployment of fenestrated stent grafts in Fenestrated Endovascular Aortic Repair (FEVAR) requires accurate geometrical alignment. Currently, this process is guided by 2D fluoroscopy which is uninformative and error prone.  In this paper, a real-time framework is proposed to instantiate the 3D shape of a fenestrated stent graft utilizing only a single low-dose 2D fluoroscopic image. Firstly, markers were placed on the fenestrated stent graft. Secondly, the 3D pose of each stent segment was instantiated by the RPnP (Robust Perspective-n-Point) method. Thirdly, the 3D shape of the whole stent graft was instantiated via graft gap interpolation. Focal U-Net was proposed to segment the markers from 2D fluoroscopic images to achieve semi-automatic marker detection. The proposed framework was validated on five patient-specific 3D printed aortic aneurysm phantoms and three stent grafts with new marker placements, showing an average distance error of $1-3 mm$ and an average angular error of $4^\circ$.
\end{abstract}

\section{INTRODUCTION}
\begin{figure}[thpb]
\centering
\framebox{\includegraphics[scale=0.48]{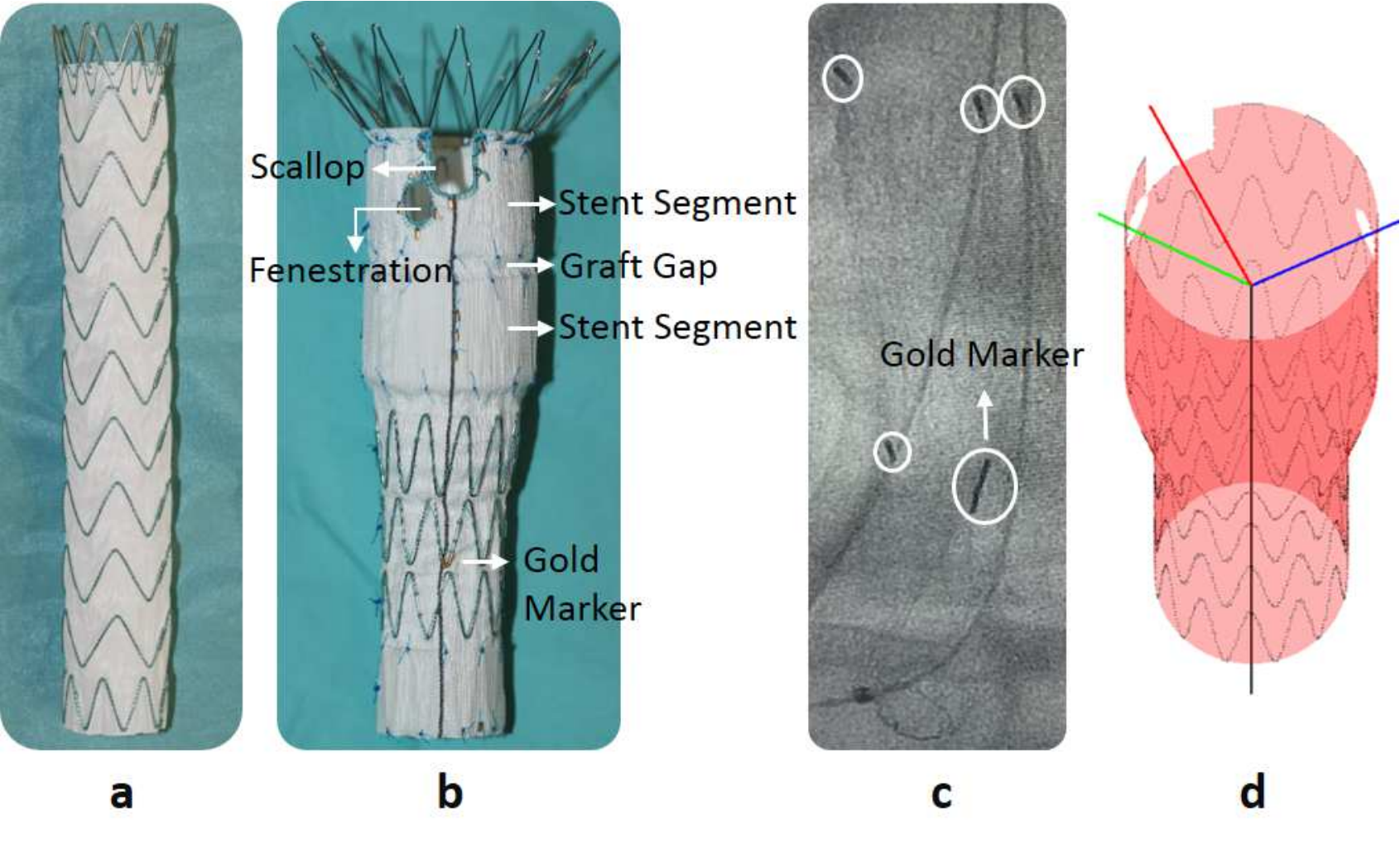}}
\caption{(a) a regular stent graft used in EVAR, (b) a fenestrated stent graft used in FEVAR with fenestrations, scallop and gold markers onside, (c) a fluoroscopic image example during FEVAR under normal radiation dose, (d) safe paths for robot-assisted vessel-fenestration cannulation. The black path is along the centrelline of the deployed main fenestrated stent graft while the green, blue and red path are from the black path end and aiming at the center of the two fenestrations and the one scallop.}
\label{fig:instruction}
\end{figure}

Endovascular Aortic Repair (EVAR), for the treatment of Abdominal Aortic Aneurysm (AAA), involves the insertion of compressed stent grafts via the femoral artery, advancement through the vasculature, subsequent device deployment, and exclusion of the aneurysmal wall. Blood flow is re-established through the deployed stent graft with reduced pressure on the diseased aneurysmal wall. The risk of rupture is abolished in the absence of endoleaks. For patients whose aneurysms involve or are adjacent to the renal and visceral vessels, Fenestrated Endovascular Aortic Repair (FEVAR) is necessary; this includes the use of a fenestrated stent graft with fenestrations or scallops to allow perfusion of vital aortic branches and ensure optimum aneurysm exclusion \cite{cross2012fenestrated}. A regular stent graft used in EVAR and a fenestrated stent graft used in FEVAR are shown in Fig. \ref{fig:instruction}a and Fig. \ref{fig:instruction}b, respectively. In addition to the location and size of fenestrations and scallops, the size and length of the stent graft are also customized according to patient-specific aortic geometries. An increasing number of stent graft manufacturers, such as Cook Medical (IN, USA) and Vascutek (Scotland, UK), are supplying fenestrated stent grafts today \cite{Endovasculartoday}.

FEVAR is a challenging and complex procedure with multiple steps. The principal challenge is the alignment of the fenestrations or scallops with the target vessels. Selective cannulation of the target vessels through the fenestrations, and subsequent branch stent graft delivery and deployment, are paramount to ensure successful aneurysm exclusion. This step can be challenging and time-consuming due to vessel tortuosity and angulation, leading to prolonged procedure and fluoroscopy time with a significant radiation burden to patients and operators \cite{cross2012fenestrated}. Alternative cannulation strategies have therefore been explored such as robotic catheter systems aiming to improve navigational accuracy and stability. One commercially available system is the Magellan (Hansen Medical, CA, USA) which includes a master-slave catheter and guidewire driving system. Clinical experience with endovascular robotic systems is growing with potential advantages of increased accuracy, safety, and stability whilst minimizing the radiation exposure \cite{riga2013robot}.

Despite advances in endovascular robotic technologies, navigation is still dependent on 2D fluoroscopy as shown in Fig. \ref{fig:instruction}c. Both the stent and graft have poor visibility under fluoroscopy. High dosage fluoroscopy may improve the visualization, however, this will increase the radiation dose. To improve FEVAR navigation, markers are sewn onto the fenestrated stent grafts to indicate the position and orientation of the fenestrations and scallops (Fig. \ref{fig:instruction}b). These markers are typically made of gold, have different shapes, and can be placed in various positions to aid in alignment of the device with the anatomy. 

There has been previous research to improve stent graft deployment. Automatic detection and tracking of stent graft delivery devices from 2D fluoroscopic images have been proposed \cite{volpi2015online}, with Frangi filtering and Robust Principal Component Analysis. Optimized stent graft sizing and placement for pulmonary artery stenosis using cylindrical affine transformation and hill climbing have also been demonstrated \cite{gundelweinautomatic}. A registration scheme combined with a semi-simultaneous optimization strategy that is to take the stent graft geometry into account was proposed to overlay 3D stent shapes onto 2D fluoroscopic images for navigation \cite{demirci20113d}. However, these methods have been demonstrated on regular off-the-shelf stent grafts for EVAR but do not take into consideration fenestrations or scallops. Renal arteries and commercial markers have been highlighted on intra-operative fluoroscopic images to aid with stent graft deployment \cite{reimlautomatic}; however, this is only in 2D and does not provide the 3D stent graft shape.

It is necessary to know exactly where fenestrations or scallops are to enable complete vessel-fenestration cannulation during FEVAR. A possible 3D navigation or robotic path is shown in Fig. \ref{fig:instruction}d. The path travels along the centreline of the deployed main fenestrated stent graft (black path in Fig. \ref{fig:instruction}d) and then is aimed at the center of corresponding fenestrations or scallops (green, blue, red path in Fig. \ref{fig:instruction}d). In order to keep a minimum radiation dose during FEVAR, we aim to use a single fluoroscopic image of several well-placed markers for 3D shape instantiation of the deployed main stent graft body. 3D shape instantiation in this paper refers to 3D shape recovery but with only a single 2D fluoroscopy projection as the input.

After being deployed into an aneurysm, the stent graft may experience twisting, bending, rotation and translation with respect to its initial straight state, making 3D instantiation of its entire shape, orientation and deformation challenging. Most of these non-rigid deformations are caused by what we term the graft gap, shown in Fig. \ref{fig:instruction}b, which is only made up of graft fabric. For the stent segments which include the metal stent and the graft attached on it, as shown in Fig. \ref{fig:instruction}b, they tend towards their initial states closely due to their relative stiffness. Thus the deformation of the whole stent graft could be split into the rigid transformations of stent segments and the non-rigid deformations of graft gaps. 

\begin{figure}[thpb]
\centering
\framebox{\includegraphics[scale=0.33]{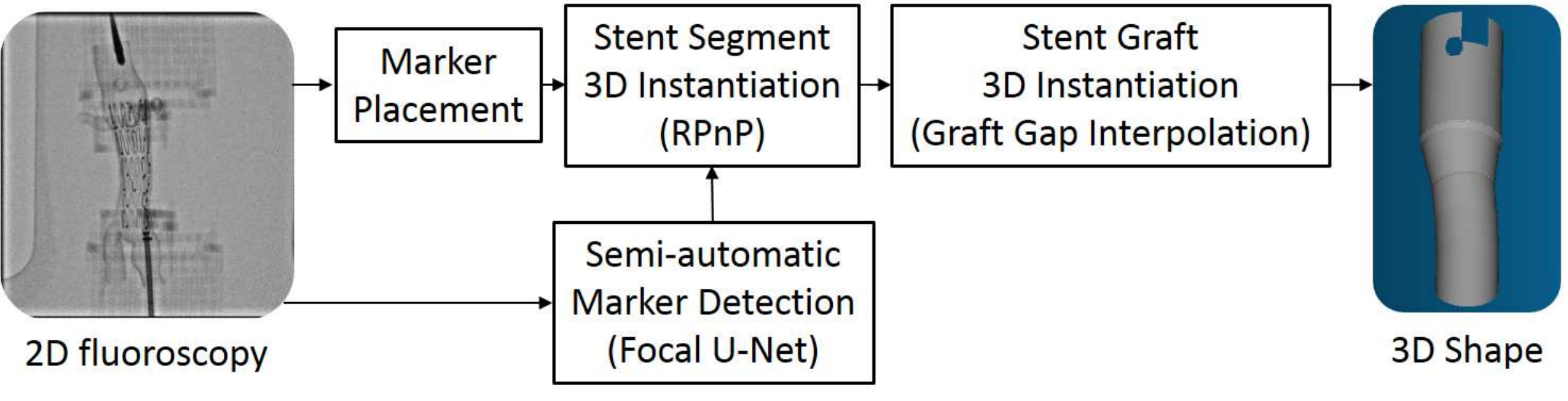}}
\caption{The proposed framework for real-time 3D shape instantiation of deployed fenestrated stent grafts.}
\label{fig:framework}
\end{figure}

We proposed a framework, as shown in Fig. \ref{fig:framework}, which instantiates the 3D shape of a fenestrated stent graft from a single 2D fluoroscopic image in real-time. Firstly, each stent segment of a fenestrated stent graft was placed on five markers at different positions. Then, the rigid transformation of individual stent segment was calculated by the RPnP (Robust Perspective-n-Point) method while the non-rigid deformations of the entire stent graft were instantiated by graft gap interpolations. Finally, Focal U-Net was proposed to achieve semi-automatic marker detection. The proposed method was validated on five 3D printed AAA phantoms and three stent grafts with newly placed markers with 78 images overall.

\section{Methodology}
Stent graft modelling, 3D stent graft shape instantiation including marker placements, rigid transformation calculations of stent segments and non-rigid deformation instantiation of the whole stent graft, semi-automatic marker detection, experimental setup, and data collection will be introduced in this section.
\subsection{Stent Graft Modelling}
Previous work, i.e. \cite{demirci20113d}, usually only focused on modelling the stents for EVAR. In FEVAR, the grafts are of equal or greater importance as fenestrations and scallops are on these grafts. Computerized Tomography (CT) could be used to acquire stent 3D shapes but not for grafts, due to the poor visibility of fabric under CT. For fenestrated stent grafts, all parameters including the height, radius, gap, etc. are known via the original stent graft design and hence mathematical modelling is available.

\begin{figure}[thpb]
\centering
\framebox{\includegraphics[scale=0.71]{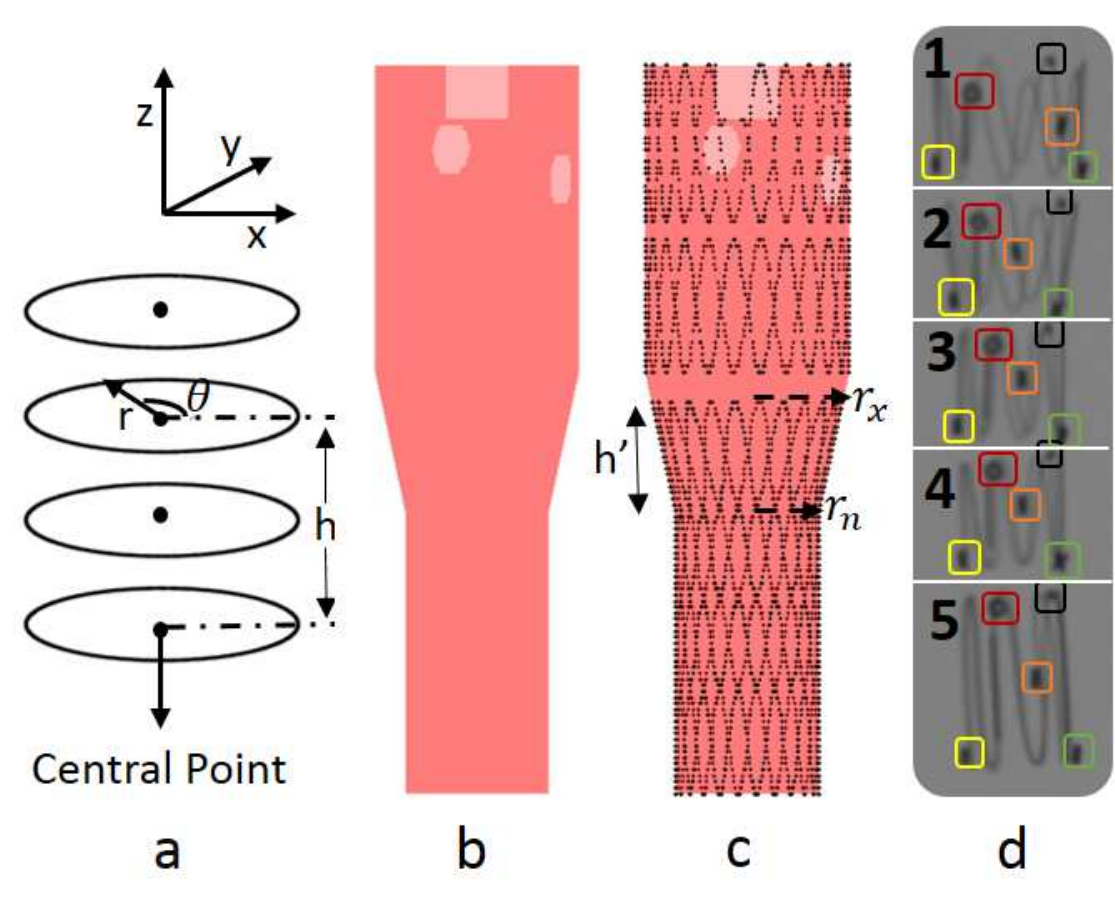}}
\caption{(a) modelling of circles, (b) modelling of graft, fenestrations and scallop, (c) modelling of a whole fenestrated stent graft, (d) marker placement and classification: markers are firstly classified into five types and then markers in each type are divided for each stent segment (five stent segments in this case).}
\label{fig:model}
\end{figure}

A stent graft is modelled with circles of different radii and set at different heights. A circle vertex was defined by $\left[\begin{array}{ccc}r*cos\theta, & r*sin\theta, & h\\ \end{array} \right]$, as shown in Fig. \ref{fig:model}a. Neighboring vertices were connected by triangles regularly to generate a surface mesh. The resolution in the height was set as $1mm$ while that in the radial direction was set as $1^\circ$ in this paper. The accumulation of these circles made up the graft modelling. To model fenestrations and scallops, vertices within the fenestration or scallop were removed (Fig. \ref{fig:model}b). $\left[\begin{array}{ccc}rcos(2\pi i/N_v), & rsin(2\pi i/N_v), & h'sin(2\pi iN_s/N_v)/2\\ \end{array} \right]$ was used to model the stent vertices \cite{demanget2012computational}, where $ r=r_n+(r_x-r_n)*(h'sin(2\pi iN_s/N_v)/2+h'/2)/h', i\in(1,N_v)$, $N_v$ is the vertex number on a stent, $N_s$ is the number of sine wave cycles describing the stent, $h'$ is the height of each stent segment (Fig. \ref{fig:model}c). For the example in Fig. \ref{fig:model}c, $r_n=11.5mm, r_x=15mm$ and $h'$ for the six stents from the bottom to top are $17mm, 13mm, 13mm, 16mm, 21mm, 25mm$ respectively. In manufacturing, stents cannot lie across fenestrations or scallops and are forced onto fenestration or scallop edges; we modelled these crossed stents onto the nearest fenestration or scallop edge too.

\subsection{3D Stent Segment Instantiation}
The non-rigid deformation of the whole stent graft was split into multiple rigid transformations of stent segments in this paper. The 3D pose of each stent segment will be instantiated based on the 2D marker fluoroscopic projections. By using the RPnP method, which estimates the pose of a calibrated camera given a set of $n$ 3D points in the world and their corresponding 2D projections in the image, the 3D pose of a stent segment could be instantiated by the 3D pose of its $n$ markers. Compared to the traditional 2D/3D registration, the RPnP method has the following advantages: 1) RPnP is fast as it solves the 3D pose mathematically; 2) RPnP is less ambiguous as it solves the 3D pose based on similar triangles; 3) RPnP only needs 4 points to instantiate a reasonable 3D pose. The correspondences between 3D points and their 2D projections are supplied by marker placement and detection in this paper.

\subsubsection{Marker Placement}
RPnP could achieve 3D pose instantiation with a minimum $n=4$ \cite{li2012robust}. We adopted $n=5$ for higher robustness. Five markers were sewn at five non-planar positions on each stent segment, as shown in Fig. \ref{fig:model}d. The marker position pattern for each stent segment is similar and is used for latter marker classification. Usually gold markers are used in commercial fenestrated stent grafts. We simulated the gold markers by printing on a Mlab Cusing R machine (ConceptLaser, Lichtenfels, Germany) with SS316L stainless steel powder. The marker size was similar to commercially used ones and the thickness guaranteed marker visibility under low radiation fluoroscopy.

\subsubsection{3D Pose Instantiation for Stent Segment}
For $n$ markers on a stent segment with known reference 3D marker positions (via the original stent graft design): $\{P_1,...,P_n\}$, after its compression and deployment, these 3D positions are transformed to target 3D marker positions: $\{P'_1,...,P'_n\}$. With known corresponding 2D marker projections (via fluoroscopy projection): $\{p_1,...,p_n\}$, the transformation matrix $\{P'_1,...,P'_n\}=Tran \cdot \{P_1,...,P_n\}$ could be instantiated by solving a RPnP problem \cite{li2012robust}.

Firstly, a rotation axis was  selected to reduce the unknown variables - here the $Z$ axis was chosen. Secondly, the PnP problem was divided into $(n-2)$ P3P problems with equation system:
\begin{equation}
f_i(x)=a_ix^4+b_ix^3+c_ix^2+d_ix+e_i = 0,i\in(1,n-2)
\label{equ:RPnP}
\end{equation}
where
\begin{equation}
\begin{array}{l}
a_i=A_6^2-A_1A_5^2\\
 b_i=2(A_3A_6-A_1A_4A_5?\\
 c_i=A_3^2+2A_6A_7-A_1A_4^2-A_2A_5^2\\
d_i=2(A_3A_7-A_2A_4A_5)\\
e_i=A_7^2-A_2A_4^2\\
A_1=k^2\\
A_2=k^2C_1^2-C_2^2\\
A_3=l_2cos\gamma_3-l_1\\
A_4=l_1cos\gamma_3-l_2\\
A_5=cos\gamma_3\\
A_6=(D_3^2-D_1^2-D_2^2)/(2D_1^2)\\
A_7=l_0^2-l_1^2-l_2^2+l_1l_2cos\gamma_3+A_6C_1^2\\
\end{array}
\label{equ:RP3P}
\end{equation}

Here $k=D_2/D_1$, $D_1,D_2,D_3$ denotes the triangle side lengths $|p_0p_1|,|p_0p_2|,|p_1p_2|$, as shown in \cite {li2011stable}. $\gamma_3=\overrightarrow{v_1} \cdot \overrightarrow{v_2}$, $C_1,C_2,l_0,l_1,l_2,\overrightarrow{v_1},\overrightarrow{v_2}$ are also as shown in \cite {li2011stable}. $x$ was solved by the local minimum of $\sum_{i=1}^{n-2}{f_i(x)}^2$. Thirdly, the depth of each marker was determined by perspective similar triangles. Fourthly, the rotation along the $Z$ axis with $c=cos\alpha,s=sin\alpha$ and translation  $\left[\begin{array}{ccc}t_x, & t_y, & t_z\\ \end{array} \right]$ of the markers were solved by \cite{li2012robust}:

\begin{equation}
\left[ \begin{array}{cc}
A_{2n\times2} & B_{2n\times4}\\
\end{array} \right]
\left[ \begin{array}{cccccc}
c & s & t_x & t_y & t_z & 1\\
\end{array} \right]^T=0
\label{equ:pose}
\end{equation}

The derivation of $A_{2n\times2}$ and $B_{2n\times4}$ was explained in \cite{li2012robust}. Finally, the solved transformation matrix was normalized by a standard 3D alignment based on Least-Squares Estimation \cite{umeyama1991least}. This normalized matrix is the 3D pose of the $n=5$ markers and the corresponding stent segment. More details of the derivation, proof, and calculation can be found in \cite{li2012robust, li2011stable} and \cite{umeyama1991least}.

\subsection{3D Stent Graft Instantiation}

\subsubsection{Continuous Constraints for Stent Segments}
\label{Sec: continuous}
In theory, the RPnP method instantiates both the position and pose accurately. In our experiments, the drifted markers, unsuitably-small delivery device and repeated stent graft compression and deployment (details explained in Sec. \ref{Sec: seg) caused non-rigid deformation between the reference and the target 3D marker positions.} When the transformation between the reference and target 3D marker positions is not rigid, errors will be added on the instantiated position and pose. The position shift of stent segments influenced the continuity of the entire stent graft and was corrected by applying continuous constraints on the circle central points.

\subsubsection{Graft Gap Interpolation}After instantiating the pose and correcting the position drift for each stent segment, the normal vectors and positions of graft gap circles were interpolated linearly by the normal vectors and positions of neighboring stent segment circles. With graft gap vertices $[r_ic_{\theta+T}, r_is_{\theta+T}, 0]$, here $T\in(1^\circ,360^\circ)$ controls the twisting and rotating of a circle, $\theta\in(1^\circ,360^\circ)$ is the angle of a vertex, $r_i$ is the radius, the interpolated graft gap vertices were calculated by:
\begin{equation}
\begin{array}{l}
\left[ \begin{array}{c}
x' \\ y' \\ z' \\ \end{array} \right] =
\left[ \begin{array}{c}
x_i \\ y_i \\ z_i\\ \end{array} \right] +
\left[ \begin{array}{c}
r_ic_{\theta+T}, r_is_{\theta+T}, 0 \end{array} \right]\cdot
\\
\\
\left[\begin{array}{ccc}
c_\Omega+\alpha ^2c_{\Omega p} & \alpha \beta c_{\Omega p}-\delta s_\Omega & \alpha \delta c_{\Omega p}+\beta s_\Omega \\
\alpha \beta c_{\Omega p}+\delta s_\Omega & c_\Omega+\beta ^2 c_{\Omega p} & \beta \delta c_{\Omega p}-\alpha s_\Omega\\
\alpha \delta c_{\Omega p}-\beta s_\Omega & \beta \delta c_{\Omega p}+\alpha s_\Omega & c_\Omega+\delta ^2c_{\Omega p}\\\end{array}\right]
\end{array}
\label{equ:model}
\end{equation}
where
\begin{equation}
c_{\Omega p} = 1-c_\Omega
\end{equation}

The rotation matrix rotates the normal vector of initial graft gap plane to be parallel to the interpolated one and was derived according to \cite{RotationMatrix}. Here, $c_{\theta+T}$ represents $cos(\theta+T)$ and $c_\Omega$ represents $cos(\Omega)$. $s_{\theta+T}$ represents $sin(\theta+T)$ and $s_\Omega$ represents $sin(\Omega)$. $\Omega$ is the angle between the circle normal vector and the xy plane (the xy plane is shown in Fig. \ref{fig:model}a). $\left[ \begin{array}{ccc}\alpha, & \beta, & \delta\\ \end{array} \right]$  controls the bending and is the cross product of the circle normal and $\left[ \begin{array}{ccc}0, & 0, & -1\\ \end{array} \right]$. $\left[ \begin{array}{ccc}x_i, & y_i, & z_i\\ \end{array} \right]$ translates the rotated graft gap vertices to the interpolated position.

\subsection{Semi-automatic Marker Detection with Deep Learning}
Semi-automatic marker detection including automatic marker segmentation and manual marker classification was proposed to find the correspondences between the 3D markers and their 2D projections. After segmenting all markers by Focal U-Net, they were classified manually into different marker types and stent segments.

\begin{figure}[thpb]
\centering
\framebox{\includegraphics[scale=0.5]{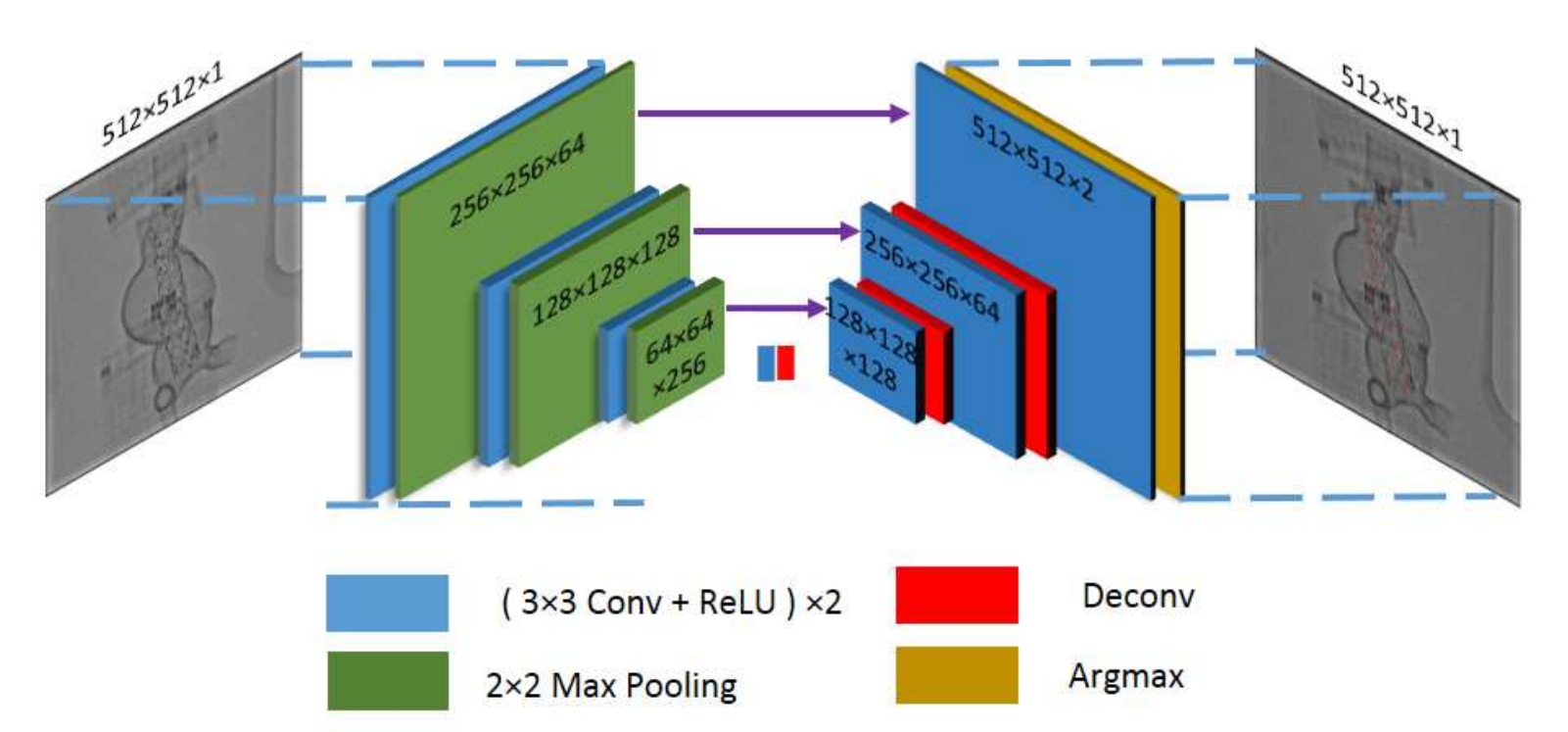}}
\caption{The network structure of Focal U-Net.}
\label{fig:unet}
\end{figure}

Focal U-Net - U-Net \cite{ronneberger2015u} trained with focal loss \cite{lin2017focal} - is proposed in this paper to segment the markers automatically from a $512\times512$ 2D fluoroscopic image. U-Net is a widely applied convolutional neural network for image segmentation. It has demonstrated high accuracies in medical problems. It manages to formulate image segmentation into pixel-level classification, hence to enable end-to-end training. The Focal U-Net structure in this paper is shown in Fig. \ref{fig:unet}. All convolutional layers were with zero padding and a stride of $1$. All max pooling layers were with a stride of $2$.

Softmax with cross-entropy loss is usually applied for no-overlap segmentation (no overlap between classes) in U-Net. For binary segmentation, it is defined as:
\begin{equation}
  loss_{cross-entropy}(p,y) =
  \begin{cases}
    -log(p) & \text{if $y = 1$} \\
    -log(1-p) & \text{if $y = 0$} \\
  \end{cases}
\label{equ:ce_loss}
\end{equation}
where $y=1$ is the foreground or marker, $y=0$ is the background, $p\in [0,1]$ is the probability of that pixel that is predicted by Focal U-Net to be the foreground. Eq. \ref{equ:ce_loss} could be rewritten as $loss_{cross-entropy}(p,y) =loss_{cross-entropy}(p_t) =-log(p_t)$, where
\begin{equation}
  p_t =
  \begin{cases}
    p & \text{if $y = 1$} \\
    1-p & \text{if $y = 0$} \\
  \end{cases}
\label{equ:P_sim}
\end{equation}

The distribution of two classes in our case - foreground/marker and background - are extremely imbalanced, as markers are small and only take approximately $0.1\%$ of the total pixels in the $512\times 512$ 2D fluoroscopic image. Weighted loss which adds much higher weights to the foreground loss while adds much lower weights to the background loss was usually applied to solve this kind of class-imbalance problem: $loss_{weight}(p_t,y) = -w_ylog(p_t)$. For convenience, we call it as Weighted U-Net in this paper. However, it under-performances in extremely and highly class-imbalanced problems, i.e. our case. So focal loss \cite{lin2017focal} was applied instead in this work:
\begin{equation}
  loss_{focal}(p_t) = -w_y(1-p_t)^\gamma log(p_t)
\label{equ:P_focal}
\end{equation}
where $\gamma$ is the pow coefficient. Empirically in our experiments, $w_y$ was set as $30$ for foreground pixels and $1$ for background pixels, $\gamma$ was set as $2$ as suggested by \cite{lin2017focal}. Focal loss is advantageous over weighted loss at a certain degree, this will be proven in Sec. \ref{Sec: seg}: instead of assigning constant weights for foreground and background pixels, focal loss separates "easy" (correctly classified) and "hard" (incorrectly classified) pixels dramatically and automatically in individual iterations, enabling the network to "gaze" at hard pixels by greatly reducing the loss contribution of easy pixels. 

To avoid unstable training and provide promising parameter initialization for Focal U-Net, the training procedure was divided into two steps: 1) the network was trained with weighted loss; 2) the model trained with Weighted U-Net was used as the initialization for the training with focal loss.

The segmented markers were manually classified into the five types based on the position pattern, as shown by the colorful bounding boxes in Fig. \ref{fig:model}d. In practice, fluoroscopic images are usually scanned in a coronal or oblique plane, which enables dividing the markers into the corresponding stent segment by their vertical positions, as shown by the white dividing lines in Fig. \ref{fig:model}d.

\subsection{Experimental Setup and Data Collection}
\subsubsection{Simulation of FEVAR}
\begin{figure}[thpb]
\centering
\framebox{\includegraphics[scale=0.4]{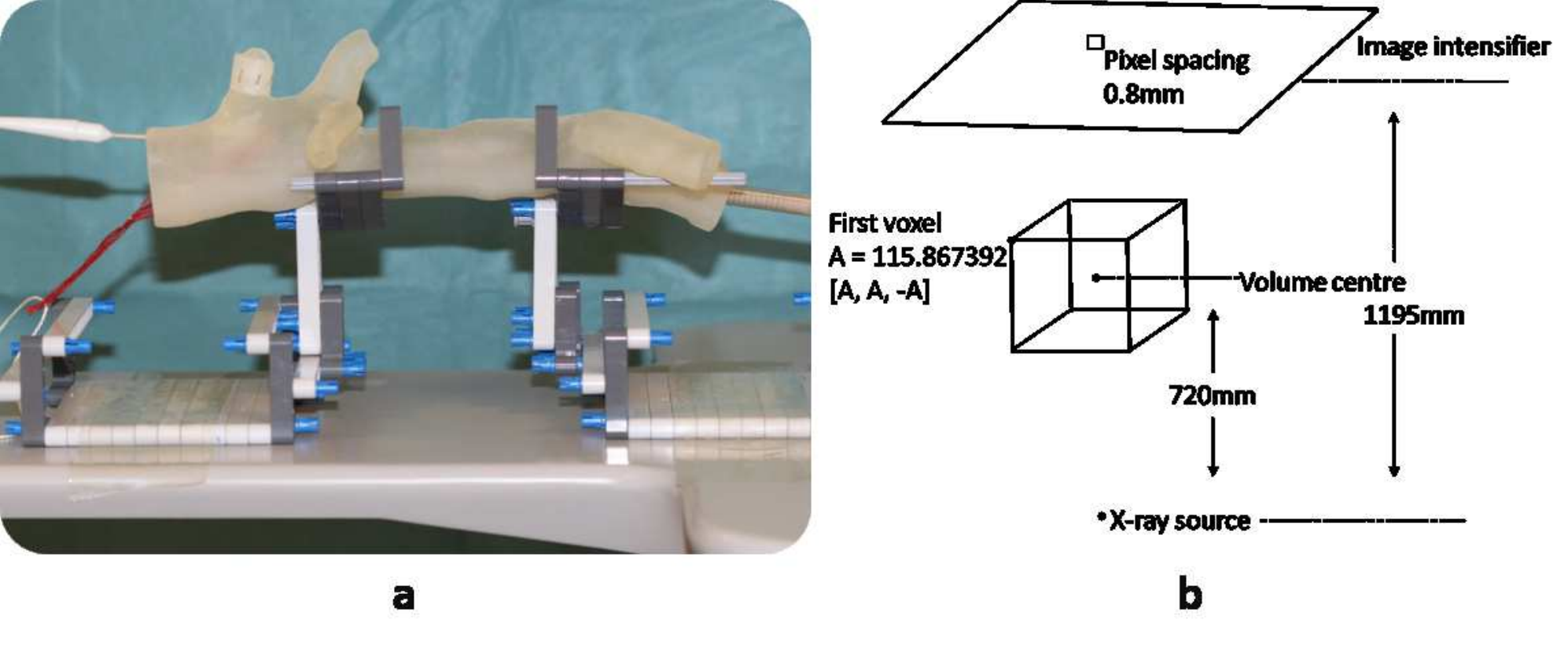}}
\caption{(a) experimental setup, (b) registration of the fluoroscopic image coordinate system to the CT coordinate system.}
\label{fig: validation}
\end{figure}

Five abdominal aneurysm phantoms, created from contrast-enhanced CT data of AAA patients, were printed on a Stratasys Objet (MN, USA) in VeroClear and TangoBlack. One example is shown in Fig. \ref{fig: validation}a. Three stent grafts: iliac ($6-10mm$ diameter, $90mm$ height, Cook Medical), fenestrated ($22-30mm$ diameter, $117mm$ height, Cook Medical) and thoracic ($30mm$ diameter, $179mm$ height, Medtronic, MN, USA) were used in the experiments. Each stent segment of the three stent grafts was sewn on five markers at non-planar positions. In a setup, a stent graft was compressed within a Captivia delivery system (Medtronic, $8mm$ diameter, shown in Fig. \ref{fig: validation}a), inserted into the 3D printed aneurysm, and deployed at the target position.

\subsubsection{Data Collection} The three stent grafts with newly sewn markers were firstly scanned by a GE Innova 4100 (GE Healthcare, Bucks, UK) for the reference 3D marker positions before any experimental setup. For simulating FEVAR, the stent graft diameter needed to fit the artery diameter, resulting in 14 matching positions in total between the five phantoms and three stent grafts. Details of each setup are shown in Tab. \ref{tab: setup}. After deploying the stent graft in each setup, 13 2D fluoroscopic images from different view angles, varying from $-90^\circ$ to $90^\circ$ with $15^\circ$ interval, were obtained by the same CT machine. There should be $14\times 13=182$ images, however, 11 images were not stored by the operator. For the setups shown in Tab. \ref{tab: setup}, $7/14$ setups expressed by $\bigodot$ were used for the training ($80$ images) of Focal U-Net, $6/14$ setups ($6\times 13=78$ images) expressed by $\bigoplus$ were used for the testing; here the test image number corresponds to that in Sec. \ref{Sec: result}, $1/14$ setup ($13$ images) expressed by $\bigotimes$ were abandoned due to one marker falling off. Sometimes, two experiments were set up with the same stent graft and the same phantom. These two setups were not the same due to the different positions inside the phantom. The $80$ training images were augmented by rotating each image from $-180^\circ$ to $165^\circ$ with $15^\circ$ intervals and flipping each rotated image along the horizontal and vertical direction, providing $5760$ training images. Each image was normalized by its maximum intensity. A CT scan was collected for each deployed stent graft. Usually, this CT scan was utilized as the ground truth of the markers and deployed stent graft, except for one comparison validation in Sec. \ref{Sec: precise} where the scanned 3D marker positions were used as the reference 3D marker positions too. The coordinates of marker projections on 2D fluoroscopic images were transformed into the CT coordinate system, as shown in Fig. \ref{fig: validation}b. The training data and ground truth for marker segmentation were labeled in Analyze (AnalyzeDirect, Inc, Overland Park, KS, USA). 3D Slicer \cite{pieper20043d} was used to segment the stent 3D shape and marker 3D shape from the CT scan. The average unsigned distance between the instantiated 3D shape and the ground truth was calculated in CloudCompare \cite{girardeau2011cloudcompare}. An additional three 2D fluoroscopic images were collected by manually generated motion of the scanning bed during fluoroscopy. This created motion-induced noise on the images, simulating the effect of respiratory and cardiac induced tissue deformation in \textit{in-vivo} scenarios.

\begin{table}
\centering
\caption{Stent Graft - Phantom Matching ($\bigoplus$ - Test; $\bigodot$ - Train; $\bigotimes$ - Abandon.)}
\begin{tabular}{|l|c|c|c|c|c|}
\hline
Phantom number & 1 & 2 & 3 & 4 & 5 \\
\hline
Iliac (S1)   & $\bigoplus$ & $\bigodot$ $\bigoplus$ &-   & $\bigodot$ & -\\ 
Test image number & $1-13$      & $14-26$                &-   & -          & -\\ 
Fenestrated (S2) & $\bigoplus$ & $\bigoplus$ &$\bigoplus$ $\bigotimes$ & $\bigodot$ $\bigodot$ & $\bigodot$\\
Test image number     & $27-39$     & $40-52$     &$53-65$                  &-                      & - \\
Thoracic (S3) & -   & $\bigodot$ &$\bigodot$   & $\bigoplus$   & -\\
Test image number  & -   &-           &-            &$66-78$        & - \\
\hline
\end{tabular}
\label{tab: setup} 
\end{table}

\section{Results}
\label{Sec: result}
Semi-automatic marker detection and 3D stent graft shape instantiation were validated with errors shown in this section. To illustrate the advantages of the proposed Focal U-Net over Weighted U-Net, an Intersection over Union (IoU) comparison of 78 images is shown in Sec. \ref{Sec: seg}. The 2D distance error of semi-automatic marker detection, the 3D distance and the angular error of marker instantiation, the 3D distance error of stent graft instantiation, instantiated 3D shape details are given in Sec. \ref{Sec: ins}, by using both manually and semi-automatically detected markers. The 3D distance error is the unsigned Euclidean distance between the instantiated 3D markers or stent grafts and the ground truth with the position displacement (explained in Sec. \ref{Sec: continuous}) corrected by aligning the centers. The angular error is the unsigned angle ($\theta$ in Fig. \ref{fig:model}a) difference between the instantiated marker and the ground truth. Angular errors were measured, as the facing and orientations of fenestrations or scallops are important for path planning (red, green, blue path in Fig. \ref{fig:instruction}d) in robot-assisted FEVAR. To evaluate the robustness of the proposed 3D stent graft shape instantiation to cardiac and respiratory induced motion, the three images with motion artifacts were tested in Sec. \ref{Sec: noise}. A comparison with using pre-experimental and intra-experimental 3D marker positions as the reference 3D marker positions is stated in Sec. \ref{Sec: precise}, showing possible accuracy improvements for \textit{in-vivo} applications. 

\subsection{2D Marker Segmentation}
\label{Sec: seg}
\begin{figure}[thpb]
\centering
\framebox{\includegraphics[scale=0.615]{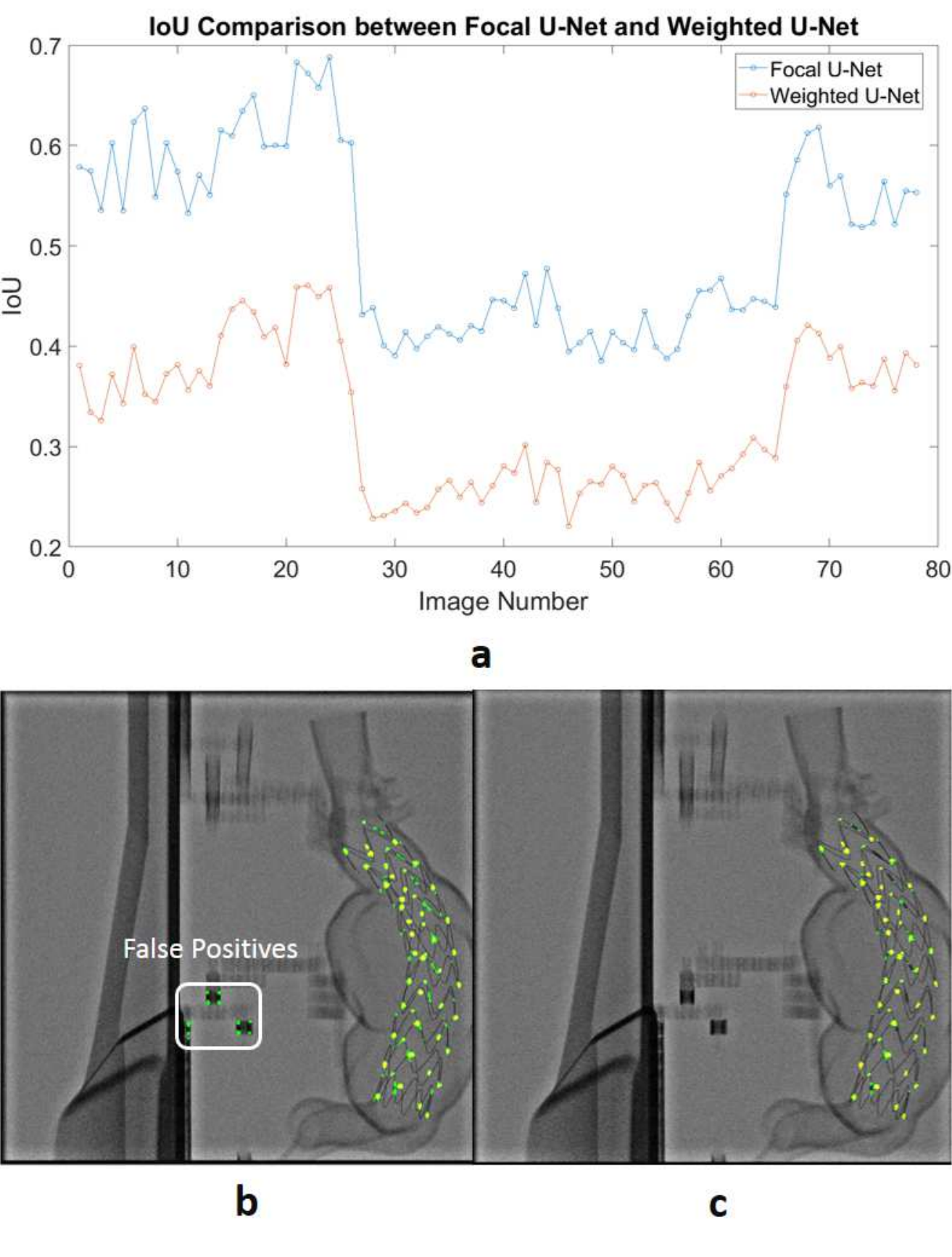}}
\caption{The segmentation IoU of Weighted U-Net and Focal U-Net for $78$ images (a); one image example segmented by Weighted U-Net (b) and Focal U-Net (c), red is the ground truth, green is the segmentation result, yellow is the overlap.}
\label{fig: seg}
\end{figure}

A total of $78$ images were segmented with the model trained by Focal U-Net and Weighted U-Net, with the IoU of each image shown in Fig. \ref{fig: seg}a. The Focal U-Net achieved a mean IoU of $0.51$ while the Weighted U-Net achieved a mean IoU of $0.33$. A segmentation example of Focal U-Net and Weighted U-Net are shown in Fig. \ref{fig: seg}b and c respectively. We can see that Focal U-Net could remove the false positives at a certain degree. The IoU for S2 (image 27-65) is less than that for S1 (image 1-26) and S3 (image 66-78), as extra commercial gold markers were placed on S2 to indicate the fenestrations and scallop. These extra markers would be segmented as well and hence decreased the IoU.

\subsection{3D Shape Instantiation}
\label{Sec: ins}
\begin{figure}[thpb]
\centering
\framebox{\includegraphics[scale=0.5]{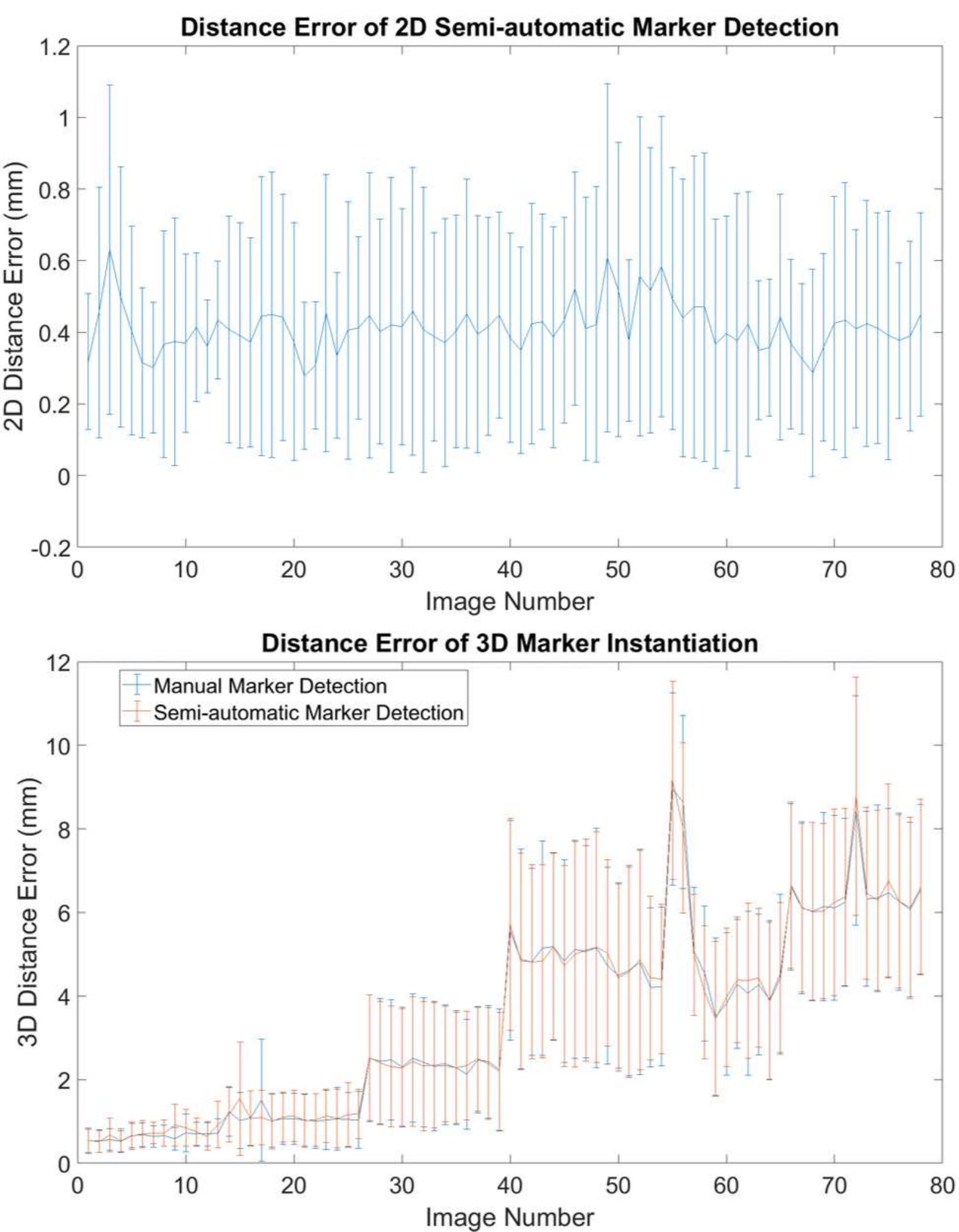}}
\caption{The (mean $\pm$ stdev) distance errors of semi-automatic marker detection (top) and 3D marker instantiation (bottom), the std errors were calculated across multiple markers on a stent graft.}
\label{fig: marker_err}
\end{figure}

The markers segmented by Focal U-Net were classified into different marker types and stent segments manually. The distance error of semi-automatic marker center determination is shown in Fig. \ref{fig: marker_err} top. An average distance error of 0.42 mm (half a pixel) was achieved. Both the marker centers detected manually and semi-automatically were used to instantiate the 3D markers. The 3D distance errors of marker instantiation are shown in Fig. \ref{fig: marker_err} bottom. An average distance error of $0.92mm$ for S1, $4.08mm$ for S2, and $6.52mm$ for S3 were achieved with semi-automatic marker detection, which were close to that achieved by manual marker detection ($0.86mm$ for S1, $4.08mm$ for S2, and $6.44mm$ for S3). This average distance error is comparable, as the marker size is almost $3mm$. The errors of S2 (image 27-65) and S3 (image 66-78) are higher than S1 (image 1-26) due to two reasons: 1) the diameters of S2 and S3 are larger than S1; 2) the deployment device was small for S2 and S3, causing more non-rigid stent segment deformations and hence more non-rigid deformations between the reference and target 3D marker positions . The errors of the latter two setups (image 40-65) are higher than that of (image 27-39), as the more times S2 was deployed, the more non-rigid stent segment deformations were added.

\begin{figure}[thpb]
\centering
\framebox{\includegraphics[scale=0.45]{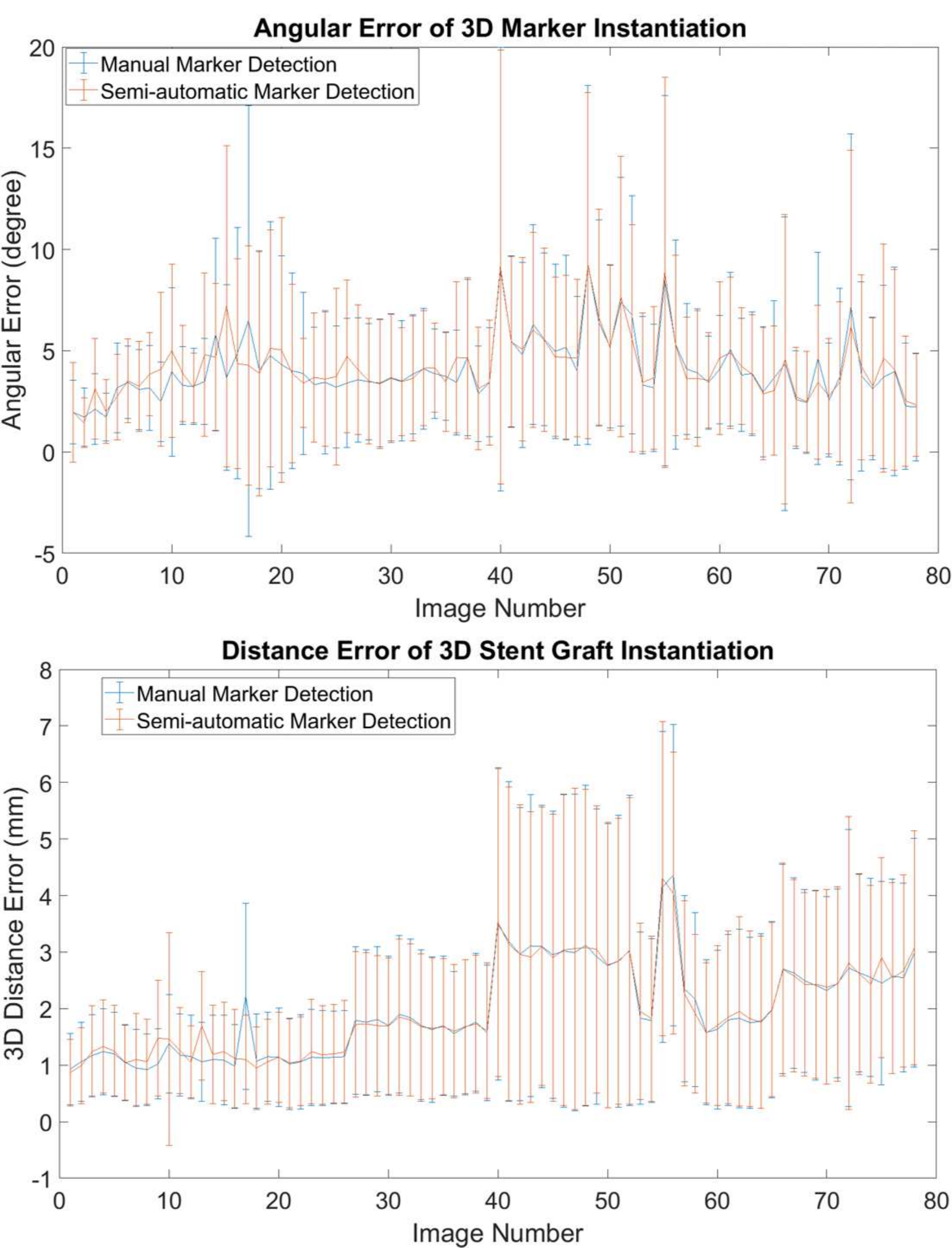}}
\caption{3D Shape instantiation errors (mean $\pm$ stdev) of angular (top) and distance (bottom) for three stent grafts. The std of angular error was calculated across multiple markers on a stent graft while that of distance error was calculated across multiple vertices of a stent graft.}
\label{fig: instantiation_error}
\end{figure}

The angular errors of instantiated markers and distance errors of instantiated stent grafts are shown in Fig. \ref{fig: instantiation_error}. An average angular error of ${4.24}^\circ$ was achieved with semi-automatic marker detection which is similar to the ${4.12}^\circ$ achieved with manual marker detection. An average distance error of $1.99mm$ was achieved with semi-automatic marker detection which is close to the $1.97mm$ achieved with manual marker detection. The average angular and distance errors for the six setups are shown in Tab. \ref{tab:markers}.

\begin{table}[h]
\caption{average errors(S1-iliac; S2-fenestrated; S3-thoracic; M-Manual; S-Semi-automatic; Angle-degree; Distance-mm)}
\label{tab:markers}
\begin{center}
\begin{tabular}{|l|c|c|c|c|c|c|c|c|}
\hline
Stent Graft    & S1   & S1   & S2   & S2   & S2   & S3\\
\hline
Image Number   & 1-13   & 14-26   & 27-39   & 40-52   & 53-65   & 66-78\\
Angle (S)      & 3.29 & 4.43 & 3.79 & 6.11 & 4.27 & 3.58\\
Angle (M)      & 2.83 & 4.25 & 3.66 & 6.18 & 4.24 & 3.57\\
Distance (S)   & 1.22 & 1.14 & 1.70 & 3.03 & 2.22 & 2.61\\
Distance (M)   & 1.10 & 1.18 & 1.72 & 3.04 & 2.23 & 2.57\\
\hline
\end{tabular}
\end{center}
\end{table}

\begin{figure}[thpb]
\centering
\framebox{\includegraphics[scale=0.36]{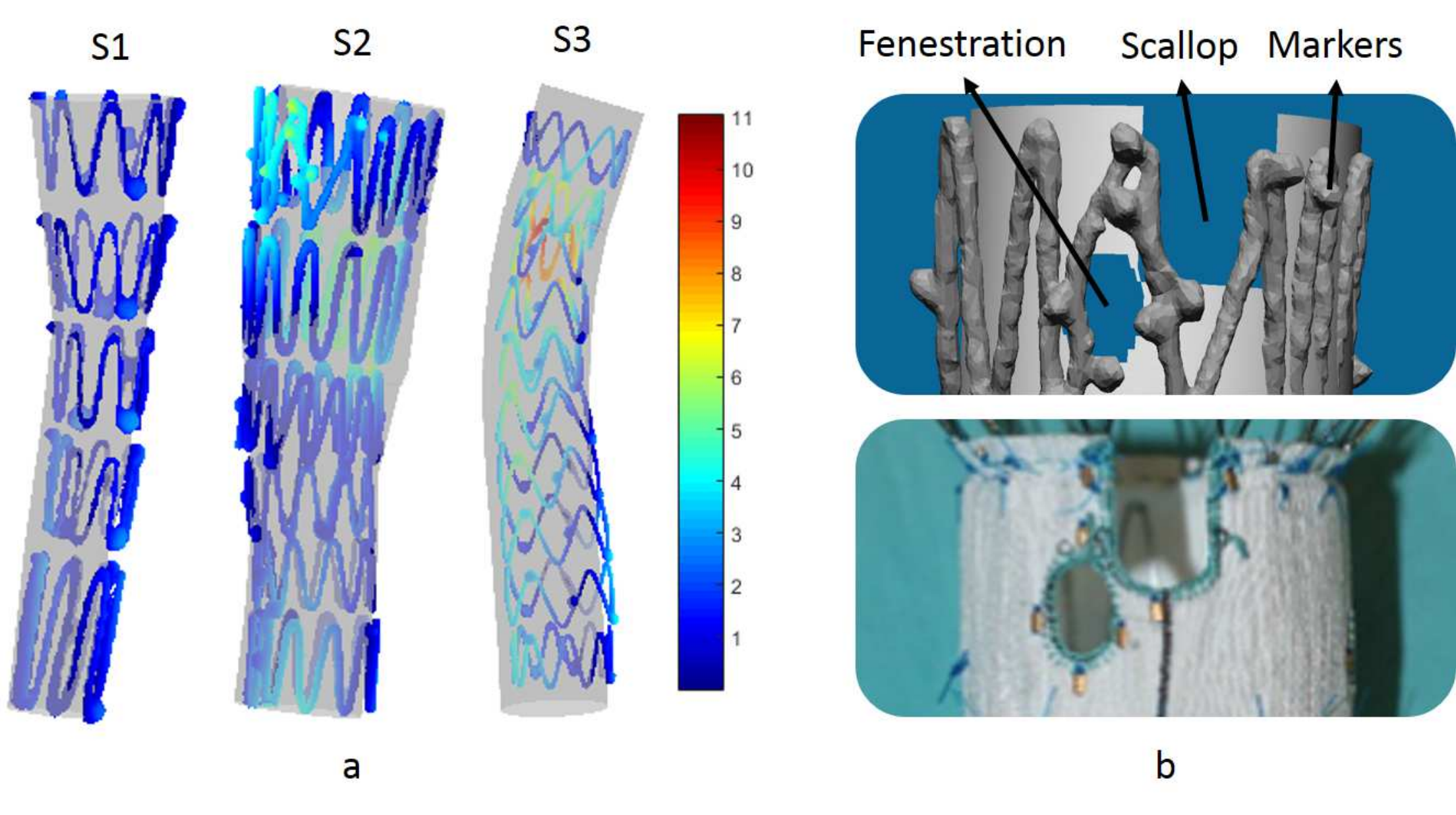}}
\caption{Examples of (a) 3D shape instantiation of the three stent grafts colored by the distance error (colorbar of errors in $mm$), (b) instantiated scallop and fenestration (top) compared to the real ones (bottom).}
\label{fig:example}
\end{figure}

Examples of 3D shape instantiation coloured by the distance error are shown in Fig. \ref{fig:example}a - the light grey mesh is the proposed shape instantiation result while the coloured stents are the ground truth. It can be seen that the bending, compressing, twisting, etc. of the stent graft, the scallops or fenestrations are instantiated well. Examples of the instantiated scallop and fenestration (Fig. \ref{fig:example}b top) are compared to the real ones (Fig. \ref{fig:example}b bottom). The dark grey stents in Fig. \ref{fig:example}b top are the ground truth from CT with commercial gold markers indicating the scallop and fenestration.

\subsection{Robustness to Motion-induced Noise}
\label{Sec: noise}
\begin{figure}[thpb]
\centering
\framebox{\includegraphics[scale=0.27]{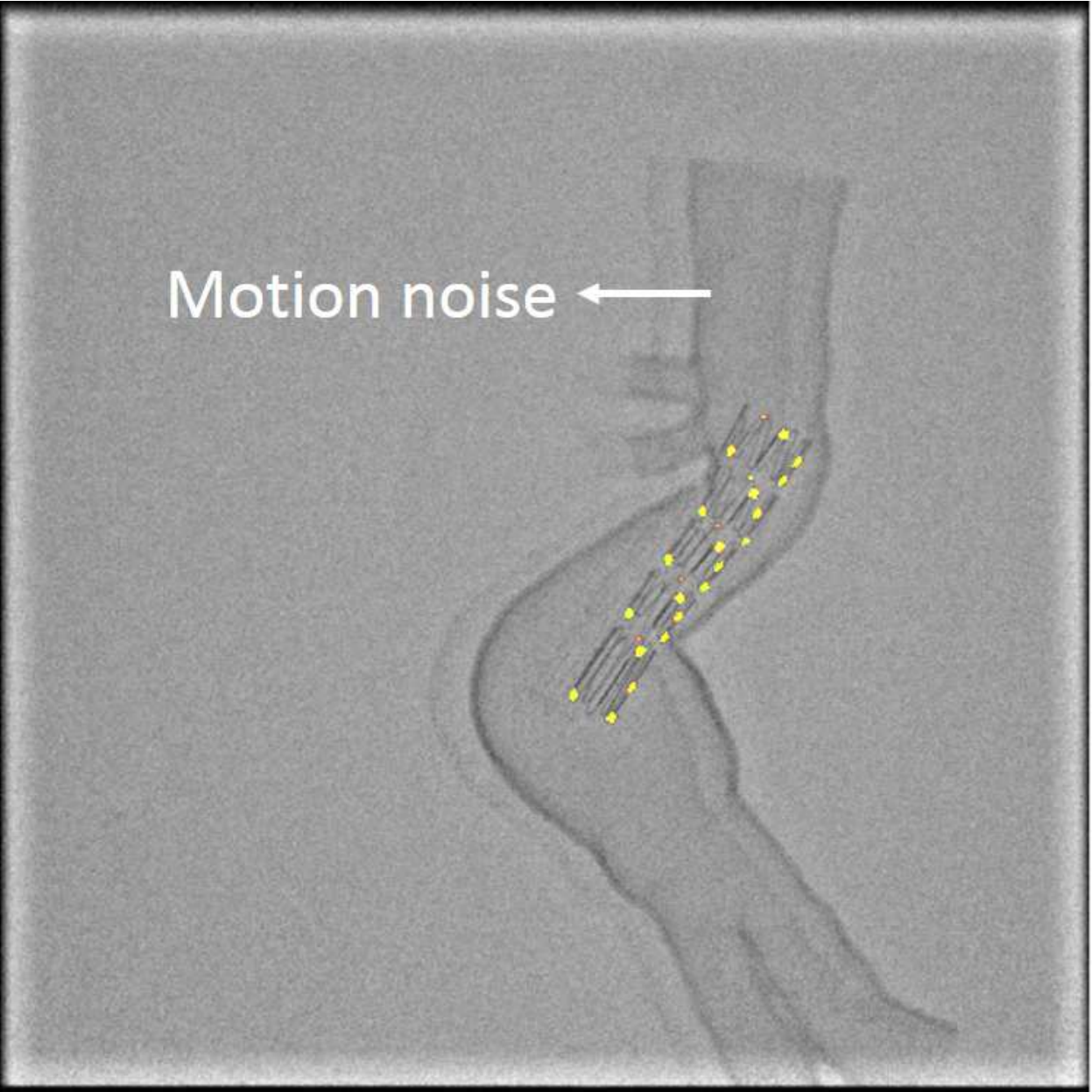}}
\caption{A segmentation example of a 2D fluoroscopic image with motion-induced noise, with the color definition the same as that in Fig. \ref{fig: seg}.}
\label{fig: motion}
\end{figure}

The three images with motion-induced noise were also evaluated and one example is shown in Fig. \ref{fig: motion}. The geometry of the aneurysm was influenced by additional artifacts while the markers remain clearly differentiable due to the high contrast of the markers. Markers were segmented with an average IoU of $0.79$. The slightly higher IoU achieved here, compared to that in Fig. \ref{fig: seg}a, is due to its clean background.

\subsection{Influence of Non-rigid Marker Set Deformation}
\label{Sec: precise}
\begin{figure}[thpb]
\centering
\framebox{\includegraphics[scale=0.35]{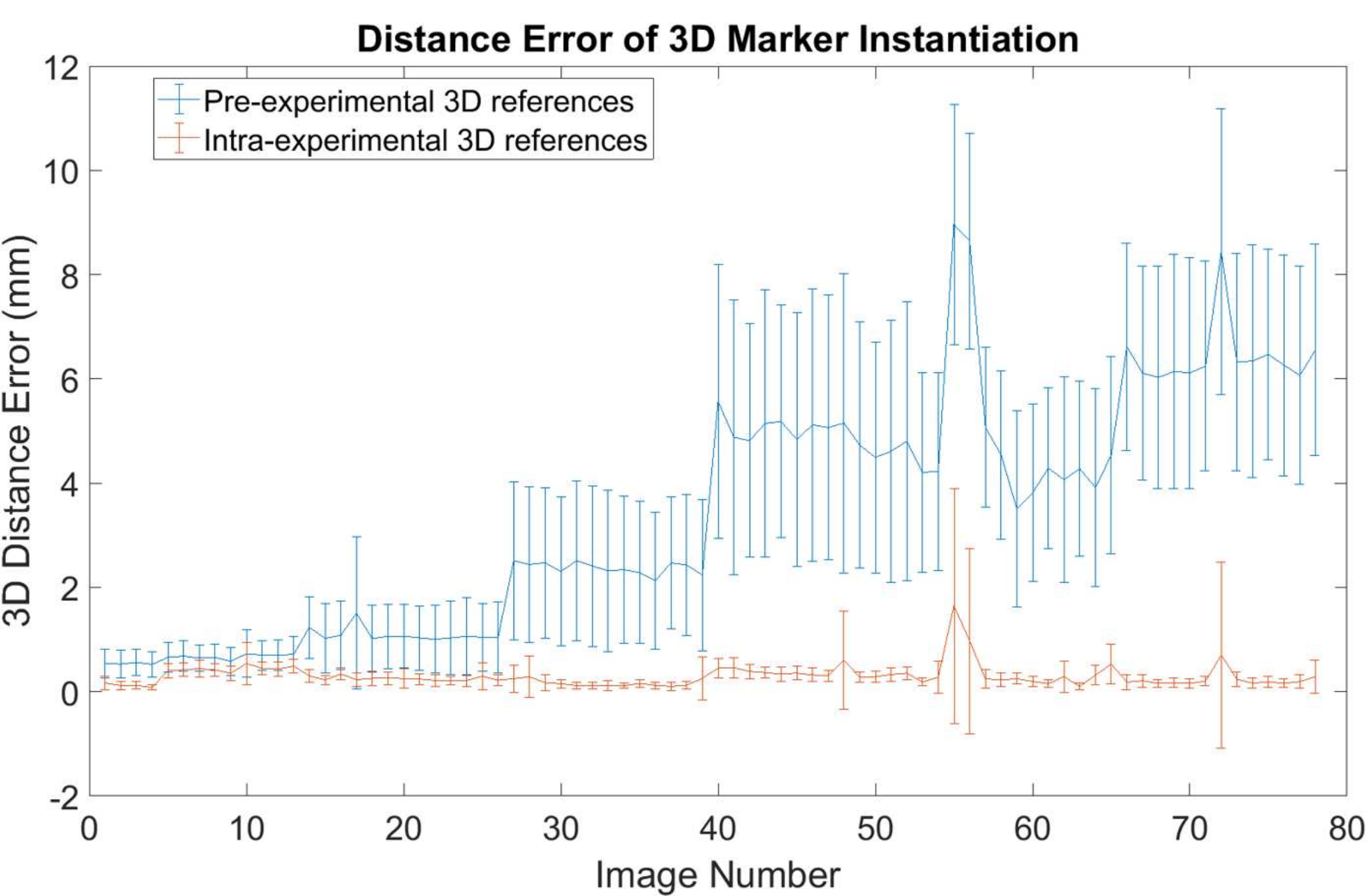}}
\caption{Distance errors of 3D marker instantiation with pre-experimental and intra-experimental 3D marker positions as the reference 3D marker positions.}
\label{fig: precise}
\end{figure}

3D marker instantiation errors with pre-experimental and intra-experimental 3D marker positions as the reference 3D marker positions are shown in Fig. \ref{fig: precise}. The errors with intra-experimental 3D references are much lower, proving that less non-rigid deformation between the reference and the target 3D marker positions could improve the instantiation accuracy. The higher errors in a few images (48, 55, 56, 72) are due to the fact that their markers were incorrectly classified.

The computational time is less than $8 ms$ in MATLAB for one stent segment instantiation on an Intel(R) Core(TM) i7-4790 CPU@3.60GHz computer. The marker segmentation takes less than $0.1s$ in Tensorflow on a NVIDIA TITAN Xp GPU. For training Focal U-Net, the first step takes about 30 minutes while the second step takes approximate 2 hours.

\section{Discussion}
In this paper, the non-rigid deformation of the whole stent graft was split into piecewise stent segment rigid transformations and then was instantiated by interpolating these instantiated stent segments. The average distance error of instantiated stent grafts at around $1-3mm$ and the average angular error of instantiated markers at around ${4}^\circ$ illustrates that this splitting is reasonable and could be used for future work on stent grafts. The average distance error of stent grafts - $3mm$ are comparable, as the size of the markers are approximately $3mm$. Even with the limited experimental environment (the drifted markers, unsuitably-small delivery device, and repeated use of the stent graft), comparable average distance and angular errors were achieved. It is expected that the accuracy could be improved with more stable marker sewing, a suitable delivery device and a one-off use of the stent graft (the stent graft is only compressed and deployed once in \textit{in-vivo scenarios}), and hence the accuracy is expected to be higher than the experiments in this paper.

The only input for the proposed 3D shape instantiation is a single fluoroscopic image of markers, which decreases the X-ray radiation to a minimum, as markers are always visible, albeit not always clearly, even under lowest X-ray radiation. Marker imaging is also robust to respiratory and cardiac induced tissue motions. The stents, 3D printed aneurysms, and the holders all show up in the 2D fluoroscopic images in our experiments - Fig. \ref{fig: seg}. In practice, the 2D fluoroscopic images, i.e. Fig. \ref{fig:instruction}c, are much more 'clean' than our experiments due to the block of tissue. The commercial markers made of gold also have higher visibility than the 3D printed markers made of steel used in this paper. It will be easier to segment and classify the markers in practical applications.

Focal U-Net was proposed in this paper, which achieved promising results in marker segmentation without any image pre-processing compared to the use of Weighted U-Net. The achieved IoU is approximately $0.51$. This is reasonable, as the number of pixels of each marker is very small; a small area of incorrectly segmented background will decrease the IoU greatly. The accuracy of segmented marker centers ($0.42mm$) and the proportion of missed markers (none was missed in all $78$ images) are more important. The markers are classified manually in this paper as the marker shapes are not well-designed into different shapes. In our test, markers with obvious different shapes - the black bounding boxes in Fig. \ref{fig:model}d - could be classified automatically by Focal U-Net. With improved marker shape design, fully automatic determination of point correspondence can be achieved.

3D shape instantiation accuracy with semi-automatic marker detection is similar to the accuracy with manual marker detection. The experiments also demonstrated the robustness of the proposed framework to fluoroscopic view angles - fluoroscopic images from 13 view angles were tested and shown with no difference in accuracy. However, a clear view with no marker overlapping and hence easier marker classification is still recommended to avoid classifying the markers incorrectly (examples shown in Sec. \ref{Sec: precise}). The running time of $0.1s$ per image indicates that the proposed framework can work in real-time; typical fluoroscopy acquisitions used in clinic are approximately 2-5 frames per second.
 
\section{Conclusion}
A 3D shape instantiation framework for fenestrated stent grafts including marker placement, stent segment pose instantiation, stent graft shape instantiation and semi-automatic marker detection  was proposed in this paper. The proposed framework only needs a single fluoroscopic image and is only based on markers, which decreases the X-ray radiation to a minimum. Compared to the current 2D fluoroscopy navigation used in robot-assisted FEVAR procedures and previous work, the proposed framework instantiates not only the 3D shapes of the stents but also the grafts, fenestrations and scallops. In the future, markers will be designed into different shapes to achieve fully automatic marker detection. This work is a first step towards a complete 3D shape instantiation which predicts the 3D shape of a fenestrated stent graft after the deployment from a single 2D fluoroscopic image of its compressed state to improve robotic navigation for FEVAR.

\section*{ACKNOWLEDGMENT}
This work was supported by EPSRC project grant EP/L020688/1. We gratefully acknowledge the support of NVIDIA Corporation with the donation of the Titan Xp GPU used for this research.

\bibliographystyle{IEEEtran}
\bibliography{ICRA2018}
\end{document}